\documentclass[conference]{IEEEtran}
\IEEEoverridecommandlockouts
\usepackage{cite}
\usepackage{amsmath,amssymb,amsfonts}
\usepackage{algorithmic}

\usepackage{textcomp}
\usepackage{xcolor}
\usepackage{xcolor}

\usepackage{graphicx}

\usepackage{array}
\usepackage{textcomp}
\usepackage{multirow}
\usepackage{booktabs}

\usepackage{mathtools}
\usepackage{breqn}
\usepackage{float}

\usepackage{fancyhdr}

\usepackage{orcidlink}
\usepackage{listings}
\usepackage{algorithmic}
\usepackage{algorithm}
\usepackage{subcaption}
\usepackage{pifont}

\usepackage{balance}

\usepackage{enumitem}

\newcommand{\xmark}{\ding{55}} 

\def\BibTeX{{\rm B\kern-.05em{\sc i\kern-.025em b}\kern-.08em
    T\kern-.1667em\lower.7ex\hbox{E}\kern-.125emX}}
\begin{document}

\title{Event-based Sensor Fusion and Application on Odometry: A Survey\\

\thanks{This research is supported by the Research Council of Finland's Digital Waters (DIWA) flagship (Grant No. 359247) as well as the DIWA Doctoral Training Pilot project funded by the Ministry of Education and Culture (Finland). }
}

\author{
    \IEEEauthorblockN{
        \vspace{1em}
        Jiaqiang Zhang\IEEEauthorrefmark{2}\,\orcidlink{0000-0002-4509-8115},
        Xianjia Yu\IEEEauthorrefmark{2}\,\orcidlink{0000-0002-9042-3730},
        Ha Sier \IEEEauthorrefmark{2}\,\orcidlink{0009-0000-3617-107X},
        Haizhou Zhang\IEEEauthorrefmark{2}\,\orcidlink{0009-0005-1321-8687},     
        Tomi Westerlund\IEEEauthorrefmark{2}\,\orcidlink{0000-0002-1793-2694}
    }
    \IEEEauthorblockA{
        \normalsize
        \IEEEauthorrefmark{2}\href{https://tiers.utu.fi}{Turku Intelligent Embedded and Robotic Systems (TIERS) Lab, University of Turku, Finland}.\\
        Emails: \textsuperscript{1}\{jiaqiang.zhang, xianjia.yu, sierha, hazhan, tovewe\}@utu.fi\\[+6pt]
    }
}

\maketitle

\begin{abstract} 
Event cameras, inspired by biological vision, are asynchronous sensors that detect changes in brightness. They offer notable advantages in environments characterized by high-speed motion, low lighting, or wide dynamic range. These distinctive properties render event cameras particularly effective for sensor fusion in robotics and computer vision, especially in enhancing traditional visual or LiDAR-inertial odometry. Conventional frame-based cameras suffer from limitations such as motion blur and drift, which can be mitigated by the continuous, low-latency data provided by event cameras. Similarly, LiDAR-based odometry encounters challenges related to the loss of geometric information in environments such as corridors. To address these limitations, unlike the existing event camera-related surveys, this survey presents a comprehensive overview of recent advancements in event-based sensor fusion for odometry applications particularly investigating fusion strategies that incorporate frame-based cameras, inertial measurement units, and LiDAR. The survey critically assesses the contributions of these fusion methods to improving odometry performance in complex environments, while highlighting key applications, and discussing the strengths, limitations, and unresolved challenges. Additionally, it offers insights into potential future research directions to advance event-based sensor fusion for next-generation odometry applications.
\end{abstract}


\section{Introduction}
Event cameras are bio-inspired vision sensors that operate fundamentally differently from traditional frame-based cameras. Instead of capturing images at fixed time intervals, event cameras detect changes in brightness asynchronously, resulting in a stream of events that encode changes in the scene~\cite{lichtsteiner200564x64, lichtsteiner2008}. This unique sensing modality provides significant advantages in scenarios involving high-speed motion, low lighting, or scenes with significant dynamic range, where traditional cameras may struggle~\cite{gallego2020event}.

\begin{figure}[t]
    \centering
    \includegraphics[width=0.48\textwidth]{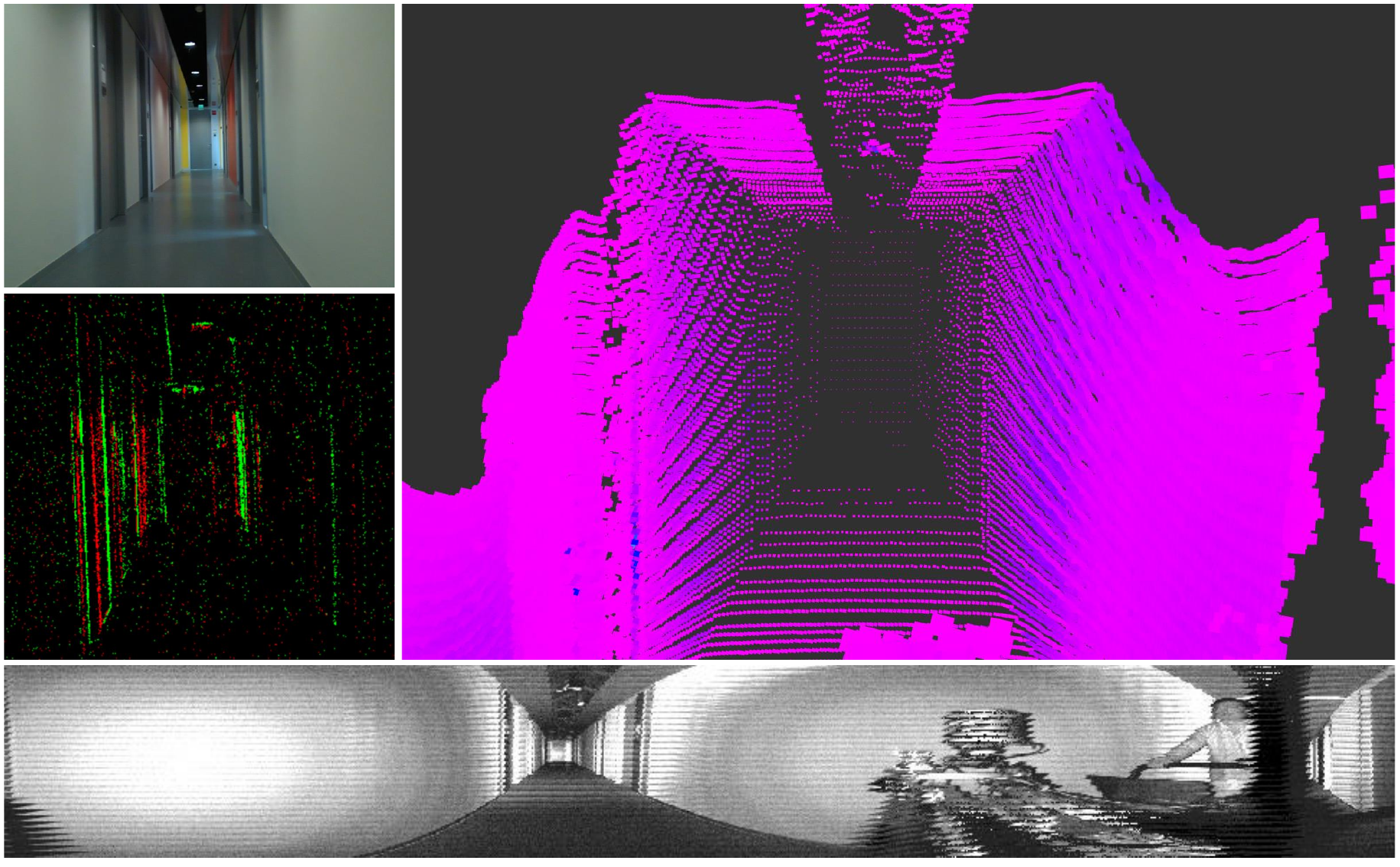}
    \caption{The sensor data representation of a long corridor includes the following: RGB image (top-left), an event-based camera image (middle-left), a LiDAR point cloud (top-right), and a LiDAR-generate reflectivity image (bottom). LiDAR data is adapted from~\cite{sier2023benchmark}.}
    \label{fig:event-rgb-lidar}
    \vspace{-1em}
\end{figure}

In the context of robotics and computer vision, sensor fusion involving event cameras has garnered considerable attention, particularly for enhancing visual odometry (VO) or visual-inertial odometry (VIO). 
Event-based sensor fusion leverages the complementary strengths of event cameras and other sensors (e.g., frame-based cameras, inertial measurement units (IMUs), LiDAR) to improve robustness and accuracy in harsh environments with high-speed needs. Regarding odometry, LiDAR-based systems generally offer greater accuracy and robustness to varying lighting conditions compared to VO or VIO. However, in environments where geometric information is homogenous, such as long corridors, LiDAR odometry is prone to drift~\cite{ha2024enhancing}. Event cameras, with their ability to capture dynamic scene changes, have the potential to mitigate this issue by providing complementary information that enhances the overall accuracy and robustness of odometry in such challenging scenarios. Fig.~\ref{fig:event-rgb-lidar} illustrates the varying representations of a long corridor as captured by different sensors, including an RGB camera, LiDAR, and an event camera. It is evident from the LiDAR data that the corridor is not clearly detected; however, by integrating the data from the RGB camera and the event camera, this limitation can be mitigated.

\begin{table*}[t]
\centering
\caption{Event Camera Products}
\label{tab:even-manufactures}
\resizebox{0.7\textwidth}{!}{
\begin{tabular}{@{}lcccl@{}}
\toprule
\textbf{Brand} & \textbf{Main Products}& \textbf{Key Features}&\textbf{Stereo Kit} \\ 
\midrule
Prophesee    & EVK4, EVK3    &  High Resolution    &    \xmark    \\ 
IniVation    & DAVIS346, DVXplorer     & Larger community     & \checkmark       \\  
Lucid    &   Triton2 EVS   &  Industrial application, PoE, GigE    &    \xmark    \\ 
Sony    & IMX636, IMX637, IMX646, IMX647     & Sensor manufacturer    &    \xmark \\   
\bottomrule
\end{tabular}
}
\end{table*}

The primary motivation for using event cameras in sensor fusion is their ability to provide continuous, low-latency information that complements the limitations of traditional sensors. For instance, frame-based cameras are prone to motion blur during rapid movements, while IMUs suffer from drift over time. By fusing event camera data with other modalities, it is possible to enhance the accuracy of visual odometry, particularly in challenging environments such as fast-changing scenes or low-light conditions.


Unlike other event-camera-based surveys~\cite{gallego2020event, shariff2024event, jia2022event, iddrisu2024event, chakravarthi2024recent}, we are particularly interested in the integration of even cameras for odometry purposes. This survey aims to provide a most comprehensive and systematic overview of recent advancements in event-based sensor fusion and its application to visual odometry. We explore different fusion strategies, including event and frame-based camera fusion, event and IMU fusion, and event and LiDAR fusion, examining their contributions to improving visual odometry performance. We also discuss various methodologies, highlight important applications, and analyze the strengths and limitations of these approaches. Additionally, we identify open challenges and promising directions for future research, providing insights into how event-based sensor fusion can be further developed to meet the needs of next-generation robotic systems.

The remainder of this paper is structured as follows: Section~\ref{sec:bg} provides an overview and fundamentals of event camera technology. Subsequently, Section~\ref{sec:SFodoR} discusses of how sensor fusion is used in robotic odometry. Next, Section~\ref{sec:fusion} explores the key trends in event camera-based sensor fusion tasks particularly on enhancing odometry by integrating event cameras with other sensors and introducing the most recent datasets relevant to event cameras for odometry. Finally, Section~\ref{sec:conclude} concludes the paper, offering insights into potential future research directions.


\begin{figure*}
    \centering
    \includegraphics[width=0.9\linewidth]{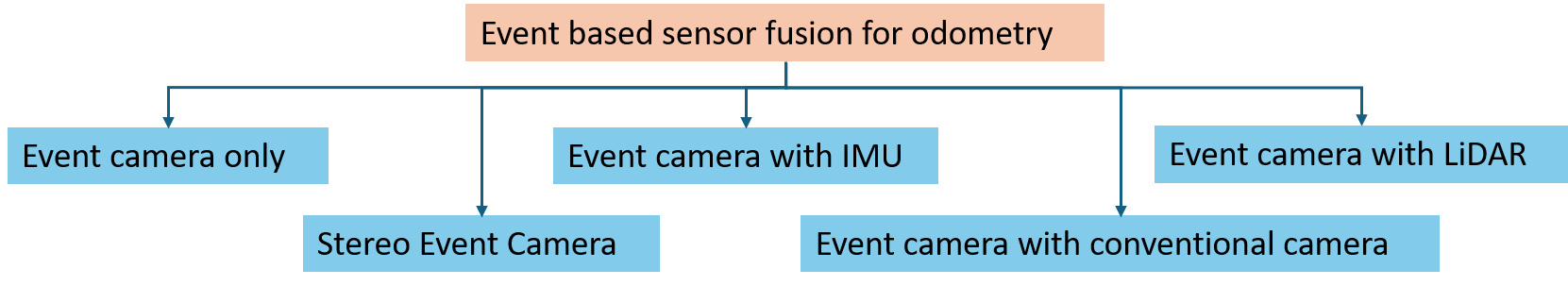}
    \caption{Primary aspects covered in event camera-based fusion for odometry purposes in this survey.}
    \label{fig:event-camera}
\end{figure*}

\section{Event Camera Technology}\label{sec:bg}

In robotics and computer vision, event cameras offer several advantages over traditional frame-based cameras, making them particularly suitable for these specific applications~\cite{gallego2020event, shariff2024event}. Currently, there are several manufacturers of event cameras, each offering different key features. Table~\ref{tab:even-manufactures} lists these manufacturers along with their main products and characteristics.

Event cameras capture changes in brightness asynchronously for each pixel. Unlike traditional frame-based cameras that capture intensity images at fixed intervals, event cameras detect changes in the logarithmic brightness of individual pixels independently, producing events whenever a change exceeds a predefined threshold~\cite{shariff2024event}. Specifically, an event at a pixel is triggered if the logarithmic intensity $\Delta L(x, y, t)$ changes since the last event surpasses the contrast threshold:
\begin{equation*}
\Delta L(x, y, t) = \log I(x, y, t) - \log I(x, y, t_{\text{last}}) = p \cdot C
\end{equation*}
\noindent where $I(x, y, t)$ is the intensity at pixel $(x, y)$ at time $t$; $t_{\text{last}}$ is the time of the last event at pixel $(x, y)$; $p \in \{+1, -1\}$ represents the polarity of the event; $C$ is the contrast threshold determining the camera’s sensitivity to brightness changes.

\subsection{Comparison with Traditional Cameras}
Event cameras capture changes in brightness asynchronously for each pixel. Unlike traditional frame-based cameras that capture intensity images at fixed frame rates, event cameras detect changes in the logarithmic brightness of individual pixels independently, producing events whenever a change exceeds a predefined threshold~\cite{shariff2024event}. Specifically, an event at a pixel is triggered if the logarithmic intensity changes since the last event surpasses the contrast threshold. This difference leads to several key advantages:
\begin{itemize}[label= {}, left=5pt]
    \renewcommand{\labelitemi}{$\ast$}
    \item \textbf{High Temporal Resolution:} Event cameras only generate data when a change occurs in the scene, resulting in sparse, storage-efficient output, unlike traditional cameras that produce dense frames at fixed intervals regardless of scene changes.
    \item \textbf{Low Latency:} Each pixel in an event camera operates independently, transmitting brightness changes immediately without global exposure, resulting in minimal latency (10\,ms in lab tests and sub-milliseconds in real-world conditions).
    \item \textbf{High Dynamic Range:} With a dynamic range of up to 140\,dB (compared to 60-70\,dB in traditional cameras), event cameras perform well in environments with significant lighting variations, avoiding saturation and underexposure.
    \item \textbf{Reduced Motion Blur:} Event cameras capture asynchronous changes at each pixel, making them inherently resistant to motion blur, especially in fast-moving or low-light conditions, ideal for visual odometry in dynamic environments.
    
\end{itemize}

Despite these advantages, event cameras also present challenges, such as the difficulty of interpreting event streams and the need for specialized algorithms to process asynchronous data. Traditional frame-based cameras, on the other hand, provide rich spatial information that is easier to use for tasks like feature extraction and object recognition, which makes them more suitable for certain applications where high spatial resolution is required.

The complementary nature of event cameras and traditional cameras has led to the development of fusion approaches that leverage the strengths of both modalities. By combining the high temporal resolution and dynamic range of event cameras with the structured spatial information from traditional cameras, it is possible to achieve more robust and accurate perception systems.

\section{Sensor Fusion for Odometry in Robotics}\label{sec:SFodoR}

Traditional VO methods can suffer from challenges such as motion blur, occlusion, and sensitivity to lighting variations. Event cameras, with their high temporal resolution and ability to capture asynchronous changes, offer a promising solution to address these challenges by providing complementary information that enhances the robustness and accuracy of VO. Using sensor fusion, we can integrate the data from these and other sensors, such as visual sensors, IMUs, LiDARs and radars, to achieve more reliable and accurate measurements. 

These sensors are used in several studies to enhance robustness of odometry estimation for robotics. Fast-LIO~\cite{xu2021fast},~\cite{bai2022faster} combined LiDAR and IMU sensors to achieve odometry, where LiDAR points were extracted as planar and edge features. An iterative Kalman filter was used to fuse these features with IMU measurements, providing accurate state estimation. Building on this, COIN-LIO~\cite{pfreundschuh2024coin} introduced an improvement by incorporating photometric information from LiDAR intensity data. The system projected LiDAR intensity values into an image format and applied filtering to enhance brightness consistency. The key features in these intensity images were selected based on their complementarity to the geometric features of the LiDAR point cloud, specifically targeting areas with limited geometric information.

Camera images, on the other hand, offer rich semantic information, making them crucial for robotic perception. To leverage this, camera data is often fused with other sensors such as IMUs (providing acceleration and angular velocity) or LiDAR (providing precise depth information). This fusion significantly improves odometry performance, particularly in complex environments or under poor lighting conditions. Frameworks like VINS-Mono~\cite{qin2018vins}, which fuses camera and IMU data, and Kimera-VIO~\cite{rosinol2020kimera, rosinol2021kimera}, which integrates LiDAR, camera, and IMU data, exemplify the advantages of such approaches. These systems enable more reliable navigation, even in low-texture or poorly lit environments, resulting in more accurate and robust odometry for real-world applications.




\section{Event-based Sensor Fusion for Odometry}\label{sec:fusion}
\subsection{Event-Camera Only}
Since an event camera only detects the change in brightness, it provides less features for localization than frame cameras; therefore, the number of studies using only an event camera for odometry is small. The first event-based pose estimation work was presented by Weikersdorfer and Conradt~\cite{weikersdorfer2012event} in 2012. This work used only an event camera and was extended to an event-based visual SLAM system in 2013~\cite{weikersdorfer2013simultaneous}.

\subsection{Fusion with Frame-based Camera}
Combining event cameras with frame-based cameras allows for high temporal resolution with more structured scene information. Liu et al.~\cite{liu2016combined} presented an approach to combine both sensor modalities for object detection and tracking. 
In~\cite{Pan_2019_CVPR}, Pan et al. proposed a method to reconstruct high frame-rate, sharp videos from a single blurry frame and event data using the event-based double integral model. The fusion of asynchronous event data with frame-based data reduces blur and improves visual quality, showcasing the complementary strengths of both modalities.
Fusing an event camera with a frame camera can overcome the limitation of each separate sensor and estimate the stereo depth. For example, Wang et al.~\cite{wang2021stereo} enabled stereo depth estimation by leveraging both event-based edge information and frame-based intensity data, offering improved disparity estimation compared to state-of-the-art stereo matching algorithms.

The fusion of an event camera and a frame camera can increase the robustness and accuracy of VO. In 2014, Censi and Scaramuzza proposed a low-latency event-based VO system~\cite{censi2014low}, which was a significant advancement in reducing the latency associated with traditional frame-based methods. This system utilized asynchronous event data to compute odometry in real time, paving the way for faster and more efficient pose estimation in dynamic environments. Building upon this, Rebecq et al.~\cite{kueng2016low} introduced low-latency VO using event-based feature tracks. This method tracked feature points based on event data, achieving reliable and low-latency odometry even in conditions where traditional frame-based methods would struggle, such as in fast motion or poor lighting.
The same group developed a 6-degree-of-freedom (DoF) event-based tracking and mapping system, which relied on geometric principles to perform both parallel tracking and mapping in real time~\cite{rebecq2016evo}. This work was instrumental in showcasing the potential of event cameras for performing full 3D tracking and mapping tasks, marking another milestone in event-based VO.

In 2022, Hidalgo-Carri\'o et al. proposed event-aided direct sparse odometry (DSO), where event data was combined with traditional DSO to improve the robustness of VO in environments with challenging lighting or rapid motion changes~\cite{Hidalgo-Carrio_2022_CVPR}. This method demonstrated the fusion of event-based and frame-based approaches in improving VO performance.

\subsection{Fusion with IMU}
IMUs provide inertial data that can be fused with events to enhance motion estimation, particularly in high-speed scenarios. IMUs can provide estimations of velocity and orientation, aiding visual odometry by reducing drift and improving robustness~\cite{scaramuzza2019visual}. In~\cite{mohamed2020towards}, Mohamed et al. explored the use of event data and IMU fusion to achieve enhanced VO performance in dynamic environments. The study proposed methods to leverage the event camera's high temporal resolution with the IMU's robust inertial data to accurately estimate motion, even in highly dynamic scenarios. 
Chamorro et al. proposed new methods for integrating event data and IMU readings to achieve ultra-fast camera pose estimates~\cite{Chamorro_2023_CVPR}. They introduced two novel fusion schemes that combine constant velocity and constant acceleration prediction models with 10\,kHz ultra-fast event-based updates and slower 1\,kHz IMU updates. The findings demonstrate superior performance at a throughput that is 100 times faster than state-of-the-art.

Integrating an IMU into VO systems significantly enhances their performance by providing additional, reliable motion information that compensates for the limitations of visual data alone. 
In 2017, Zhu et al. extended the capabilities of event-based odometry by integrating inertial measurements in their work on event-based visual-inertial odometry~\cite{Zhu_2017_CVPR}. The combination of event cameras and inertial sensors allowed for robust pose estimation even in challenging scenarios, such as high-speed motion or complex dynamic scenes.
In 2018, Mueggler and his team presented a continuous-time visual-inertial odometry approach for event cameras, further enhancing the accuracy and robustness of event-based odometry systems~\cite{mueggler2018continuous}. Their work highlighted the importance of fusing temporal information from both event data and inertial sensors in a continuous-time framework.

Mahlknecht et al.~\cite{mahlknecht2022exploring} explored event camera-based odometry for planetary robots in 2022, showcasing the potential of event cameras in extraterrestrial environments where traditional cameras may struggle due to harsh conditions.
In 2023, Lee et al.~\cite{lee2023event} introduced a hybrid approach combining event and frame-based visual-inertial odometry with adaptive filtering techniques based on 8-DoF warping uncertainty. This method enhanced the odometry system's resilience to noise and dynamic changes, further improving the robustness and adaptability of event-based systems in various application settings.

\subsection{Event-based Stereo Visual Odometry}
Event-based stereo visual odometry is an evolving field that leverages the advantages of stereo event cameras, which provide asynchronous data from two synchronized event sensors, allowing for 3D depth perception in challenging environments with high-speed motion or varying lighting conditions.
The early works on Event-based stereo VO were published in 2018. Rebecq et al.~\cite{rebecq2018emvs} introduced Event-based Multi-View Stereo VO, which enabled 3D reconstruction in real-time using event cameras. This method capitalized on the asynchronous nature of events to compute depth information by fusing event data from multiple viewpoints, enabling accurate and efficient 3D reconstruction. 
In the same year, a semi-dense 3D reconstruction method using stereo event cameras was proposed~\cite{zhou2018semi}. Their approach utilized stereo event cameras to reconstruct semi-dense 3D maps, significantly improving the accuracy and robustness of visual odometry systems, even in dynamic environments.


In 2021, Zhou et al.~\cite{zhou2021event} extended their prior work in~\cite{zhou2018semi} by introducing a complete event-based stereo visual odometry system. Their method combined stereo event cameras and a robust odometry pipeline to estimate camera motion in real time. By leveraging the high temporal resolution of event cameras, this system was able to track fast-moving objects and scenes with challenging lighting conditions.
The feature-based method can also be utilized on event-based stereo visual odometry~\cite{hadviger2021feature, hadviger2023stereo}. Their approach detected and tracked features directly from stereo event data, using them to estimate motion. 

Recently, research in stereo event-based VIO has gained attention, as it enhances motion estimation by fusing inertial measurements with visual data to improve accuracy and robustness, particularly in conditions where visual information alone may be unreliable.
In 2023, Wang et al. developed a stereo event-based VIO system, which incorporated both stereo event cameras and inertial measurements to improve the accuracy of pose estimation~\cite{wang2023stereo}.
Further research into stereo event-based VIO led to the work of Chen et al.~\cite{chen2023esvio}. 
In parallel, Liu et al. also presented an event-based stereo visual-inertial odometry system~\cite{liu2023esvio}.
Their methods demonstrated superior performance in both accuracy and robustness when compared to frame-based VIO systems, particularly in scenarios involving rapid motion.

Most recent works by Niu et al. have continued to push the boundaries of event-based stereo VO. Their IMU-aided event-based stereo visual odometry system\cite{niu2024imu} and ESVO2~\cite{niu2024esvo2} incorporate direct visual-inertial odometry with stereo event cameras, significantly improving real-time pose estimation accuracy. 

\begin{table*}[t]
\centering
\caption{Event camera-based dataset related to odometry }
\label{tab:event-odom-dataset}
\resizebox{\textwidth}{!}{
\begin{tabular}{@{}lcll@{}}
\toprule
\textbf{Year} & \textbf{Dataset Name}& \textbf{Scenarios}&\textbf{Sensors} \\ 
\midrule
2024    &  CoSEC~\cite{peng2024cosec}    &  Driving scenarios     &  Event Camera, LiDAR, RGB Camera, IMU, GNSS      \\ 
2023 & ECMD~\cite{chen2023ecmd}    &  Driving scenarios     &  Event Camera, LiDAR, RGB Camera, Infrared Camera, IMU, GNSS      \\ 
2023 & UNIZG-FER LAMOR~\cite{hadviger2023stereo}    &  Indoor handheld \& Outdoor Driving    & Stereo Event Camera, LiDAR, IMU, GNSS      \\ 
2022    &  Evimo2~\cite{burner2022evimo2}    &    Indoor Scenes   & Stereo Event Camera, IMU       \\  
2021    & DSEC~\cite{gehrig2021dsec}    &   Driving scenarios   &   Stereo Event Camera, LiDAR, RGB Camera, IMU, GNSS    \\ 
2018    & MVSEC~\cite{zhu2018multivehicle}     & Handheld, UAV \& Driving     &   Stereo Event Camera, LiDAR, VI-Sensor, IMU, GNSS \\    
2017    & DAVIS Datasets~\cite{mueggler2017event}& High-speed robotics &Event Camera, RGB Camera, IMU \\
\bottomrule
\end{tabular}
}
\end{table*}

\subsection{Fusion with LiDARs}
LiDAR technology generates precise 3D maps and is capable of producing highly accurate odometry information. However, it operates at a lower frequency and encounters challenges in specific scenarios, as previously mentioned, where geometric information is compromised or lost. Several studies have focused on the fusion of event cameras with LiDAR systems, aiming to leverage the advantages of event cameras to enhance LiDAR performance.

Some of the research is still in the early stages, with a primary focus on achieving automatic and accurate calibration.
In~\cite{10161220}, Ta et al. introduced a direct extrinsic calibration method between event cameras and LiDAR, exploiting the high temporal resolution and dynamic range of event cameras. This technique eliminates the need for frame-based intermediaries and optimizes 6-DoF extrinsic calibration through information-based correlation methods, improving sensor alignment. Similarly, Song et al.~\cite{8584215} proposed a novel calibration method using a 3D marker to generate stable patterns detectable in both 2D point sets and 3D point clouds. This approach facilitates accurate calibration between the event camera and LiDAR by leveraging light changes to trigger event detection, which is crucial for achieving stable fusion. This paves the way for the fusion of event cameras and LiDARs.  

Beyond calibration, LiDAR and event camera fusion can enhance depth perception. Cui et al.~\cite{9686362} proposed a method to fuse sparse LiDAR and event data to generate dense depth maps, improving spatial perception. This improved point cloud map can further facilitate the LiDAR-based odometry. In~\cite{zhou2024bring}, Zhou et al. discovered that data from event cameras share a homogeneous nature with RGB and LiDAR data in both visual and motion spaces. This characteristic makes event cameras particularly suitable for bridging the gap between RGB and LiDAR in motion fusion or scene flow tasks. The finding highlights the potential for a hierarchical fusion framework that effectively integrates these sensors, making the fusion of event cameras with other sensors like LiDAR and conventional cameras highly promising.



\subsection{Event Camera Datasets}\label{sec:dataset}
With the increasing interest towars event cameras and their utilisation in robotic odometry, many datasets has been introduced in recent years. In Table~\ref{tab:event-odom-dataset},s we list those datasets that have particularly been developed to support research in event-based sensor fusion for odometry, enabling accurate pose estimation and motion tracking in dynamic environments.

\section{Conclusion and Discussion}\label{sec:conclude}
Due to their high temporal resolution, low latency, high dynamic range, and reduced motion blur compared to traditional cameras, event cameras offer significant potential benefits for odometry calculation by capturing precise changes in brightness. Unlike previous review papers focused on event cameras, this survey provides a comprehensive and detailed analysis of the application of event cameras for odometry, exploring various sensor fusion approaches. These include event camera-only systems, fusion with IMUs, stereo event camera setups, integration with traditional cameras, and combinations with LiDAR. Additionally, we briefly discuss event camera-related datasets that are relevant for advancing sensor fusion, particularly in odometry applications.

Event camera holds significant potential for future applications across various domains. Its role in odometry or pose estimation has been specifically addressed in the paper. In Visual Odometry (VO) or Visual-Inertial Odometry (VIO), the use of event cameras can effectively mitigate motion blur and reduce the challenges posed by low-light environments, thanks to their wider dynamic range. For LiDAR-based odometry, similar to their application with conventional cameras, event cameras can help reduce drift by compensating for the loss of geometric information in LiDAR point clouds with photometric information. 
Additionally, event cameras can be integrated with LiDAR-generated images, as modern LiDAR systems often produce dense, low-resolution panoramic images with encoded infrared light reflectivity and intensity as shown in the bottom of Fig.~\ref{fig:event-rgb-lidar}. This integration could further enhance the robustness and accuracy of odometry in complex environments. Although the current stage of fusion with LiDAR is still relatively early, it shows promising potential for future developments.


\bibliographystyle{unsrt}
\balance
\bibliography{bibliography}

\begin{thebibliography}{10}

\bibitem{lichtsteiner200564x64}
Patrick Lichtsteiner and Tobi Delbruck.
\newblock A 64x64 aer logarithmic temporal derivative silicon retina.
\newblock In {\em Research in Microelectronics and Electronics, 2005 PhD}, volume~2. IEEE, 2005.

\bibitem{lichtsteiner2008}
Patrick Lichtsteiner, Christoph Posch, and Tobi Delbruck.
\newblock A 128$\times$ 128 120 db 15 $\mu$s latency asynchronous temporal contrast vision sensor.
\newblock {\em IEEE Journal of Solid-State Circuits}, 43(2), 2008.

\bibitem{gallego2020event}
Guillermo Gallego, Tobi Delbrück, Garrick Orchard, Chiara Bartolozzi, Brian Taba, Andrea Censi, Stefan Leutenegger, Andrew~J. Davison, J{\"o}rg Conradt, Kostas Daniilidis, and Davide Scaramuzza.
\newblock Event-based vision: A survey.
\newblock {\em IEEE transactions on pattern analysis and machine intelligence}, 44(1), 2020.

\bibitem{ha2024enhancing}
Sier Ha, Honghao Du, Xianjia Yu, Jian Song, and Tomi Westerlund.
\newblock Enhancing the reliability of lidar point cloud sampling: A colorization and super-resolution approach based on lidar-generated images.
\newblock {\em arXiv preprint arXiv:2409.11532}, 2024.

\bibitem{sier2023benchmark}
Ha~Sier, Qingqing Li, Xianjia Yu, Jorge~Peña Queralta, Zhuo Zou, and Tomi Westerlund.
\newblock A benchmark for multi-modal lidar slam with ground truth in gnss-denied environments.
\newblock {\em Remote Sensing}, 15(13), 2023.

\bibitem{shariff2024event}
Waseem Shariff, Mehdi~Sefidgar Dilmaghani, Paul Kielty, Mohamed Moustafa, Joe Lemley, and Peter Corcoran.
\newblock Event cameras in automotive sensing: A review.
\newblock {\em IEEE Access}, 2024.

\bibitem{jia2022event}
Siqi Jia.
\newblock Event camera survey and extension application to semantic segmentation.
\newblock In {\em Proc. of the 4th Int. Conference on Image Processing and Machine Vision}, 2022.

\bibitem{iddrisu2024event}
Khadija Iddrisu, Waseem Shariff, Peter Corcoran, Noel O’Connor, Joe Lemley, and Suzanne Little.
\newblock Event camera based eye motion analysis: A survey.
\newblock {\em IEEE Access}, 2024.

\bibitem{chakravarthi2024recent}
Bharatesh Chakravarthi, Aayush~Atul Verma, Kostas Daniilidis, Cornelia Fermuller, and Yezhou Yang.
\newblock Recent event camera innovations: A survey.
\newblock {\em arXiv preprint arXiv:2408.13627}, 2024.

\bibitem{xu2021fast}
Wei Xu and Fu~Zhang.
\newblock Fast-lio: A fast, robust lidar-inertial odometry package by tightly-coupled iterated kalman filter.
\newblock {\em IEEE Robotics and Automation Letters}, 6(2), 2021.

\bibitem{bai2022faster}
Chunge Bai, Tao Xiao, Yajie Chen, Haoqian Wang, Fang Zhang, and Xiang Gao.
\newblock Faster-lio: Lightweight tightly coupled lidar-inertial odometry using parallel sparse incremental voxels.
\newblock {\em IEEE Robotics and Automation Letters}, 7(2), 2022.

\bibitem{pfreundschuh2024coin}
Patrick Pfreundschuh, Helen Oleynikova, Cesar Cadena, Roland Siegwart, and Olov Andersson.
\newblock Coin-lio: Complementary intensity-augmented lidar inertial odometry.
\newblock In {\em IEEE Int. Conference on Robotics and Automation (ICRA)}. IEEE, 2024.

\bibitem{qin2018vins}
Tong Qin, Peiliang Li, and Shaojie Shen.
\newblock Vins-mono: A robust and versatile monocular visual-inertial state estimator.
\newblock {\em IEEE transactions on robotics}, 34(4), 2018.

\bibitem{rosinol2020kimera}
Antoni Rosinol, Marcus Abate, Yun Chang, and Luca Carlone.
\newblock Kimera: an open-source library for real-time metric-semantic localization and mapping.
\newblock In {\em 2020 IEEE Int. Conference on Robotics and Automation (ICRA)}. IEEE, 2020.

\bibitem{rosinol2021kimera}
Antoni Rosinol, Andrew Violette, Marcus Abate, Nathan Hughes, Yun Chang, Jingnan Shi, Arjun Gupta, and Luca Carlone.
\newblock Kimera: From slam to spatial perception with 3d dynamic scene graphs.
\newblock {\em The Int. Journal of Robotics Research}, 40(12-14), 2021.

\bibitem{weikersdorfer2012event}
David Weikersdorfer and J{\"o}rg Conradt.
\newblock Event-based particle filtering for robot self-localization.
\newblock In {\em 2012 IEEE Int. Conference on Robotics and Biomimetics (ROBIO)}. IEEE, 2012.

\bibitem{weikersdorfer2013simultaneous}
David Weikersdorfer, Raoul Hoffmann, and J{\"o}rg Conradt.
\newblock Simultaneous localization and mapping for event-based vision systems.
\newblock In {\em Computer Vision Systems}, Berlin, Heidelberg, 2013. Springer Berlin Heidelberg.

\bibitem{liu2016combined}
Hongjie Liu, Diederik~Paul Moeys, Gautham Das, Daniel Neil, Shih-Chii Liu, and Tobi Delbr{\"u}ck.
\newblock Combined frame-and event-based detection and tracking.
\newblock In {\em 2016 IEEE Int. Symposium on Circuits and systems (ISCAS)}. IEEE, 2016.

\bibitem{Pan_2019_CVPR}
Liyuan Pan, Cedric Scheerlinck, Xin Yu, Richard Hartley, Miaomiao Liu, and Yuchao Dai.
\newblock Bringing a blurry frame alive at high frame-rate with an event camera.
\newblock In {\em Proc. of the IEEE/CVF Conference on Computer Vision and Pattern Recognition (CVPR)}, June 2019.

\bibitem{wang2021stereo}
Ziwei Wang, Liyuan Pan, Yonhon Ng, Zheyu Zhuang, and Robert Mahony.
\newblock Stereo hybrid event-frame (shef) cameras for 3d perception.
\newblock In {\em 2021 IEEE/RSJ Int. Conference on Intelligent Robots and Systems (IROS)}. IEEE, 2021.

\bibitem{censi2014low}
Andrea Censi and Davide Scaramuzza.
\newblock Low-latency event-based visual odometry.
\newblock In {\em 2014 IEEE Int. Conference on Robotics and Automation (ICRA)}. IEEE, 2014.

\bibitem{kueng2016low}
Beat Kueng, Elias Mueggler, Guillermo Gallego, and Davide Scaramuzza.
\newblock Low-latency visual odometry using event-based feature tracks.
\newblock In {\em 2016 IEEE/RSJ Int. Conference on Intelligent Robots and Systems (IROS)}. IEEE, 2016.

\bibitem{rebecq2016evo}
Henri Rebecq, Timo Horstsch{\"a}fer, Guillermo Gallego, and Davide Scaramuzza.
\newblock Evo: A geometric approach to event-based 6-dof parallel tracking and mapping in real time.
\newblock {\em IEEE Robotics and Automation Letters}, 2(2), 2016.

\bibitem{Hidalgo-Carrio_2022_CVPR}
Javier Hidalgo-Carri\'o, Guillermo Gallego, and Davide Scaramuzza.
\newblock Event-aided direct sparse odometry.
\newblock In {\em Proc. of the IEEE/CVF Conference on Computer Vision and Pattern Recognition (CVPR)}, June 2022.

\bibitem{scaramuzza2019visual}
Davide Scaramuzza and Zichao Zhang.
\newblock Visual-inertial odometry of aerial robots.
\newblock {\em arXiv preprint arXiv:1906.03289}, 2019.

\bibitem{mohamed2020towards}
Sherif~AS Mohamed, Mohammad-Hashem Haghbayan, Mohammed Rabah, Jukka Heikkonen, Hannu Tenhunen, and Juha Plosila.
\newblock Towards dynamic monocular visual odometry based on an event camera and imu sensor.
\newblock In {\em Intelligent Transport Systems. From Research and Development to the Market Uptake: Third EAI Int. Conference, INTSYS 2019}. Springer, 2020.

\bibitem{Chamorro_2023_CVPR}
William Chamorro, Joan Sol\`a, and Juan Andrade-Cetto.
\newblock Event-imu fusion strategies for faster-than-imu estimation throughput.
\newblock In {\em Proc. of the IEEE/CVF Conference on Computer Vision and Pattern Recognition (CVPR) Workshops}, June 2023.

\bibitem{Zhu_2017_CVPR}
Alex Zihao~Zhu, Nikolay Atanasov, and Kostas Daniilidis.
\newblock Event-based visual inertial odometry.
\newblock In {\em Proc. of the IEEE Conference on Computer Vision and Pattern Recognition (CVPR)}, July 2017.

\bibitem{mueggler2018continuous}
Elias Mueggler, Guillermo Gallego, Henri Rebecq, and Davide Scaramuzza.
\newblock Continuous-time visual-inertial odometry for event cameras.
\newblock {\em IEEE Transactions on Robotics}, 34(6), 2018.

\bibitem{mahlknecht2022exploring}
Florian Mahlknecht, Daniel Gehrig, Jeremy Nash, Friedrich~M Rockenbauer, Benjamin Morrell, Jeff Delaune, and Davide Scaramuzza.
\newblock Exploring event camera-based odometry for planetary robots.
\newblock {\em IEEE Robotics and Automation Letters}, 7(4), 2022.

\bibitem{lee2023event}
Min~Seok Lee, Jae~Hyung Jung, Ye~Jun Kim, and Chan~Gook Park.
\newblock Event-and frame-based visual-inertial odometry with adaptive filtering based on 8-dof warping uncertainty.
\newblock {\em IEEE Robotics and Automation Letters}, 2023.

\bibitem{rebecq2018emvs}
Henri Rebecq, Guillermo Gallego, Elias Mueggler, and Davide Scaramuzza.
\newblock Emvs: Event-based multi-view stereo—3d reconstruction with an event camera in real-time.
\newblock {\em Int. Journal of Computer Vision}, 126(12), 2018.

\bibitem{zhou2018semi}
Yi~Zhou, Guillermo Gallego, Henri Rebecq, Laurent Kneip, Hongdong Li, and Davide Scaramuzza.
\newblock Semi-dense 3d reconstruction with a stereo event camera.
\newblock In {\em Proc. of the European conference on computer vision (ECCV)}, 2018.

\bibitem{zhou2021event}
Yi~Zhou, Guillermo Gallego, and Shaojie Shen.
\newblock Event-based stereo visual odometry.
\newblock {\em IEEE Transactions on Robotics}, 37(5), 2021.

\bibitem{hadviger2021feature}
Antea Hadviger, Igor Cvi{\v{s}}i{\'c}, Ivan Markovi{\'c}, Sacha Vra{\v{z}}i{\'c}, and Ivan Petrovi{\'c}.
\newblock Feature-based event stereo visual odometry.
\newblock In {\em 2021 European Conference on Mobile Robots (ECMR)}. IEEE, 2021.

\bibitem{hadviger2023stereo}
Antea Hadviger, Vlaho-Josip {\v{S}}tironja, Igor Cvi{\v{s}}i{\'c}, Ivan Markovi{\'c}, Sacha Vra{\v{z}}i{\'c}, and Ivan Petrovi{\'c}.
\newblock Stereo visual localization dataset featuring event cameras.
\newblock In {\em 2023 European Conference on Mobile Robots (ECMR)}. IEEE, 2023.

\bibitem{wang2023stereo}
Kunfeng Wang, Kaichun Zhao, and Zheng You.
\newblock Stereo event-based visual-inertial odometry.
\newblock {\em arXiv preprint arXiv:2303.05086}, 2023.

\bibitem{chen2023esvio}
Peiyu Chen, Weipeng Guan, and Peng Lu.
\newblock Esvio: Event-based stereo visual inertial odometry.
\newblock {\em IEEE Robotics and Automation Letters}, 8(6), 2023.

\bibitem{liu2023esvio}
Zhe Liu, Dianxi Shi, Ruihao Li, and Shaowu Yang.
\newblock Esvio: event-based stereo visual-inertial odometry.
\newblock {\em Sensors}, 23(4), 2023.

\bibitem{niu2024imu}
Junkai Niu, Sheng Zhong, and Yi~Zhou.
\newblock Imu-aided event-based stereo visual odometry.
\newblock {\em arXiv preprint arXiv:2405.04071}, 2024.

\bibitem{niu2024esvo2}
Junkai Niu, Sheng Zhong, Xiuyuan Lu, Shaojie Shen, Guillermo Gallego, and Yi~Zhou.
\newblock Esvo2: Direct visual-inertial odometry with stereo event cameras.
\newblock {\em arXiv preprint arXiv:2410.09374}, 2024.

\bibitem{peng2024cosec}
Shihan Peng, Hanyu Zhou, Hao Dong, Zhiwei Shi, Haoyue Liu, Yuxing Duan, Yi~Chang, and Luxin Yan.
\newblock Cosec: A coaxial stereo event camera dataset for autonomous driving.
\newblock {\em arXiv preprint arXiv:2408.08500}, 2024.

\bibitem{chen2023ecmd}
Peiyu Chen, Weipeng Guan, Feng Huang, Yihan Zhong, Weisong Wen, Li-Ta Hsu, and Peng Lu.
\newblock Ecmd: An event-centric multisensory driving dataset for slam.
\newblock {\em IEEE Transactions on Intelligent Vehicles}, 2023.

\bibitem{burner2022evimo2}
Levi Burner, Anton Mitrokhin, Cornelia Ferm{\"u}ller, and Yiannis Aloimonos.
\newblock Evimo2: an event camera dataset for motion segmentation, optical flow, structure from motion, and visual inertial odometry in indoor scenes with monocular or stereo algorithms.
\newblock {\em arXiv preprint arXiv:2205.03467}, 2022.

\bibitem{gehrig2021dsec}
Mathias Gehrig, Willem Aarents, Daniel Gehrig, and Davide Scaramuzza.
\newblock Dsec: A stereo event camera dataset for driving scenarios.
\newblock {\em IEEE Robotics and Automation Letters}, 6(3), 2021.

\bibitem{zhu2018multivehicle}
Alex~Zihao Zhu, Dinesh Thakur, Tolga {\"O}zaslan, Bernd Pfrommer, Vijay Kumar, and Kostas Daniilidis.
\newblock The multivehicle stereo event camera dataset: An event camera dataset for 3d perception.
\newblock {\em IEEE Robotics and Automation Letters}, 3(3), 2018.

\bibitem{mueggler2017event}
Elias Mueggler, Henri Rebecq, Guillermo Gallego, Tobi Delbruck, and Davide Scaramuzza.
\newblock The event-camera dataset and simulator: Event-based data for pose estimation, visual odometry, and slam.
\newblock {\em The Int. Journal of Robotics Research}, 36(2), 2017.

\bibitem{10161220}
Kevin Ta, David Bruggemann, Tim Brödermann, Christos Sakaridis, and Luc Van~Gool.
\newblock L2e: Lasers to events for 6-dof extrinsic calibration of lidars and event cameras.
\newblock In {\em 2023 IEEE Int. Conference on Robotics and Automation (ICRA)}, 2023.

\bibitem{8584215}
Rihui Song, Zhihua Jiang, Yanghao Li, Yunxiao Shan, and Kai Huang.
\newblock Calibration of event-based camera and 3d lidar.
\newblock In {\em 2018 WRC Symposium on Advanced Robotics and Automation (WRC SARA)}, 2018.

\bibitem{9686362}
Mingyue Cui, Yuzhang Zhu, Yechang Liu, Yunchao Liu, Gang Chen, and Kai Huang.
\newblock Dense depth-map estimation based on fusion of event camera and sparse lidar.
\newblock {\em IEEE Transactions on Instrumentation and Measurement}, 71, 2022.

\bibitem{zhou2024bring}
Hanyu Zhou, Yi~Chang, and Zhiwei Shi.
\newblock Bring event into rgb and lidar: Hierarchical visual-motion fusion for scene flow.
\newblock In {\em Proc. of the IEEE/CVF Conference on Computer Vision and Pattern Recognition}, 2024.

\end{thebibliography}


\end{document}